\documentclass[journal]{IEEEtran}
\usepackage{amsmath,amssymb} 
\usepackage{color}
\usepackage{multirow}
\usepackage[switch]{lineno}
\usepackage{cite}

\ifCLASSINFOpdf
  \usepackage[pdftex]{graphicx}
  \graphicspath{{../pdf/}{../jpeg/}}
  \DeclareGraphicsExtensions{.pdf,.jpeg,.png}
\else
  \usepackage[dvips]{graphicx}
  \graphicspath{{../eps/}}
  \DeclareGraphicsExtensions{.eps}
\fi
\begin{document}
%
\title{Prototype-guided Cross-task Knowledge Distillation for Large-scale Models}
%
%
%

\author{Deng Li,
        Aming Wu,
        Yahong Han, 
        Qi Tian
\thanks{Deng Li and Yahong Han are with the College of Intelligence and Computing, Tianjin University, Tianjin, China. (email: lideng@tju.edu.cn; yahong@tju.edu.cn)}
\thanks{Aming Wu is with the School of Electronic Engineering, Xidian University, Xi'an, China. (email: amwu@xidian.edu.cn).}
\thanks{Qi Tian is with Cloud \& AI, Huawei Technologies, Shenzhen, China (email: tian.qi1@huawei.com).}
}

%
%

\markboth{Journal of \LaTeX\ Class Files,~Vol.~XX, No.~XX, August~XX}%
{Shell \MakeLowercase{\textit{et al.}}: Prototype-guided Cross-task Knowledge Distillation for Large-scale Models}
%



\maketitle

\begin{abstract}
Recently, large-scale pre-trained models have shown their advantages in many tasks. However, due to the huge computational complexity and storage requirements, it is challenging to apply the large-scale model to real scenes. A common solution is knowledge distillation which regards the large-scale model as a teacher model and helps to train a small student model to obtain a competitive performance. Standard knowledge distillation methods mainly require the teacher model and the student model to perform the same task, \emph{e.g.}, the student model and the teacher model share the same label space, which limits its application in the real scenario. Cross-task Knowledge distillation can transfer the knowledge of teacher models to student models of different label spaces without fine-tuning, which expands the application scenarios of the large-scale pre-trained model. Existing knowledge distillation works focus on directly mimicking the final prediction or the intermediate layers of the teacher model, which represent the global-level characteristics and are task-specific. To alleviate the constraint of different label spaces, capturing invariant intrinsic local object characteristics (such as the shape characteristics of the leg and tail of the cattle and horse) plays a key role. Considering the complexity and variability of real scene tasks, we propose a Prototype-guided Cross-task Knowledge Distillation (ProC-KD) approach to transfer the intrinsic local-level object knowledge of a large-scale teacher network to various task scenarios. First, to better transfer the generalized knowledge in the teacher model in cross-task scenarios, we propose a prototype learning module to learn from the essential feature representation of objects in the teacher model. Secondly, for diverse downstream tasks, we propose a task-adaptive feature augmentation module to enhance the features of the student model with the learned generalization prototype features and guide the training of the student model to improve its generalization ability. The experimental results on various visual tasks demonstrate the effectiveness of our approach for large-scale model cross-task knowledge distillation scenes. 
 
\end{abstract}


%
\IEEEpeerreviewmaketitle

\section{Introduction}
%
%
%
%
\IEEEPARstart{R}{ecently}, the Transformer network \cite{vaswani2017attention} has achieved great advances in some visual tasks, such as image classification \cite{dosovitskiy2020image,touvron2021deit,liu2021swin}, object detection \cite{zhu2021deformable,carion2020detr}, image segmentation \cite{ye2019cross}, action recognition \cite{9834306}, and visual language joint learning \cite{tan2019lxmert,su2020vl,li2020hero,chen2020uniter}. 
Transformer networks based on a self-attention mechanism can process complete input sequences and 
own the advantage of parallelization. Therefore, it is usually used to obtain the pre-trained model from large-scale datasets \cite{deng2009imagenet}. 
Currently,  the common strategy for using large-scale pre-trained models on cross-task downstream tasks is fine-tuning. After learning the generalized feature representation from large-scale datasets, it is then fine-tuning on the downstream task with a small number of datasets to improve the performance of the small model. 
However, due to the huge computational complexity and huge storage requirements of these models, it has become a great challenge to apply them to practical application scenarios with limited resources, \emph{e.g.}, mobile devices.

To solve the above model application issue, some model compression and acceleration technologies are proposed,  \emph{e.g.}, parameter pruning \cite{zhu2017to,molchanov2016pruning}, model quantization \cite{wu2016quantized}, and knowledge distillation (KD) \cite{hinton2015distilling}. Particularly, knowledge distillation is an effective method of model compression, which distills the knowledge from a large deep neural network (teacher model) into a small network (student model) \cite{hinton2015distilling, 9729452, 9169852, 9240973, 9827964}. 
Unlike other model compression methods, knowledge distillation can reduce the size of the network and improve the performance of small models on downstream tasks regardless of the structural differences between teacher and student networks. Its success has been witnessed in a wide range of applications such as computer vision \cite{hinton2015distilling,romero2015fitnets,huang2017like,muller2019when,yim2017fps,park2019relational,gou2021knowledge}, speech recognition \cite{chebotar2016distilling,kurata2020knowledge,yoon2021tutornet}, and natural language processing \cite{sanh2019distilbert,sun2019patient,jiao2020tinybert}.  

\begin{figure*}[t]
\begin{center}
\includegraphics[width=18.5cm]{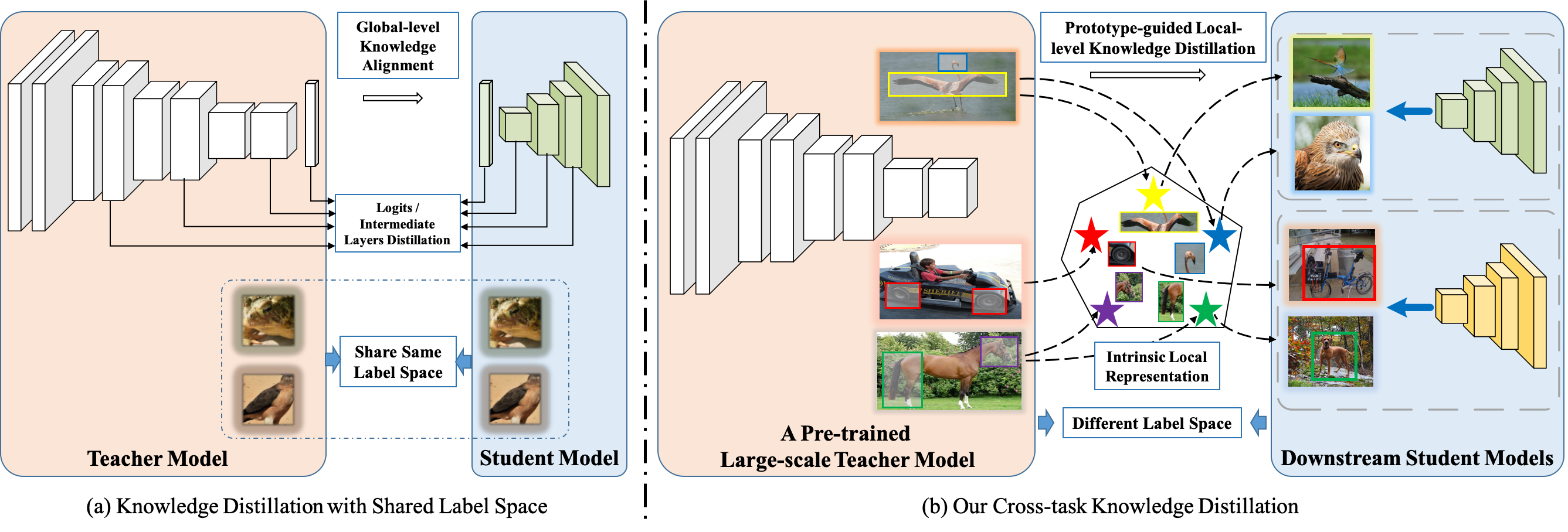}
\end{center}
   \caption{Comparison between the conventional and proposed methods. (a) The conventional knowledge distillation method in which the teacher model and student model share the same label space. The downstream task is limited to the same task as the teacher model and is global-level knowledge alignment. (b) The proposed prototype-guided cross-task knowledge distillation method in which the teacher model and student model with different label spaces. The prototypes learn the invariant intrinsic local-level representation from the embedding of the large-scale teacher model and guide the learning of various cross-task student models.}
\label{fig_intro}
\end{figure*}

However, these knowledge distillation approaches mainly require the teacher model and the student model to be the same task, \emph{e.g.}, the student model and teacher model share the same label space, which limits their application in the real scenario such as downstream tasks in different label spaces (as shown in Fig. \ref{fig_intro} (a)). 

Cross-task knowledge distillation can transfer the knowledge of the teacher model to downstream tasks in different label spaces, which expands the application of the teacher model on a variety of downstream tasks. 
Existing Same-task knowledge distillation works mainly to transfer the final prediction logits or the intermediate-layers knowledge of the teacher model, which are global-level knowledge alignments and can not be applied to cross-task knowledge distillation directly. 
Earlier cross-task knowledge distillation work \cite{ye2020distilling} aligns the high-order comparison relationship between models in a local manner, while, this method lag in the representation power of the invariant intrinsic object and is a two-stage distillation method.

Under the context of cross-task knowledge distillation, the intrinsic object characteristics can give benefit guidance to the training of the student model, for example, the shape features of the legs of a cattle and a horse when the cattle and horse belong to the datasets of the teacher model and the student model, respectively. 
Considering the complexity and variability of real scene tasks and the generalization capability of large-scale pre-trained models, in this paper, we propose a Prototype-guided Cross-task Knowledge Distillation (ProC-KD) approach to transfer the local intrinsic knowledge of a large-scale teacher network to various task scenarios (as shown in Fig. \ref{fig_intro} (b)). And our method of obtaining the downstream tasks small model is a one-stage training process.

Specifically, our proposed approach consists of two integrated modules: a prototype-based representation learning module and a feature augmentation module. 

The prototype learning module is carefully designed to capture essential feature information from the intermediate feature of the teacher model. Next, the learned prototype representation is fed into the feature augmentation module. The feature augmentation module targets at enriching the student model feature which is more related to the prototype representation while suppressing the unrelated feature.

To guide the training of the student model with the learned generalized prototype representation, we employ a consistency loss to enable the maximum agreement between the prototype augmented features and student network features.

In the experiments, we first verify the effectiveness of the proposed method on various cross-task knowledge distillation tasks. Then, we evaluate the proposed method on standard knowledge distillation tasks. The experiment results in two scenarios demonstrate the effectiveness and generality of our method.

Our contributions in this paper are summarized as follows:

(1) We propose a prototype-guided knowledge distillation approach to transfer the intrinsic knowledge from a large-scale model to different cross-task small models without fine-tuning the teacher model on the downstream task dataset and improve the student model generalization ability.

(2) We propose a prototype learning module and a feature augmentation module to learn the invariant intrinsic knowledge of the large-scale teacher model and enhance the small student model with the attention mechanism, respectively.

(3) We verify our method on both cross-task knowledge distillation and standard knowledge distillation on various visual tasks. The experiment results demonstrate the effectiveness and generality of our approach.

\section{Related Work}

\subsection{Large-scale Model}

Transformer-based large-scale models have achieved great success in natural language processing  \cite{devlin2018bert}, computer vision \cite{dosovitskiy2020image,touvron2021deit,liu2021swin}, and multi-modal task learning \cite{tan2019lxmert,su2020vl}. 

In this paper, we mainly discuss the recent efforts of the Transformer-based model in the field of computer vision. As a pioneered work, ViT \cite{dosovitskiy2020image} constructs the token embedding for the Transformer by directly dividing each image into 16 $\times$ 16 patches and projecting them into embeddings. The experiments are carried out on large-scale training datasets (\emph{e.g.}, ImageNet-21k and JFT-300M). Deit \cite{touvron2021deit} introduced several training strategies to speed up the training of ViT. 
Deit \cite{touvron2021deit} introduced a distillation method to transfer CNN-based features to a visual Transformer. However, it is not the paradigm of distilling the knowledge networks to smaller student networks, which means the parameters of a model after distillation is not greatly reduced. Swin Transformer \cite{liu2021swin} introduces shifted non-overlapping window partitions and restricts self-attention computation within sub-windows.  
Apart from Transformer-based large-scale model, series of CNN-based large-scale models \cite{radosavovic2020designing} have also been proposed. 
Radosavovic \emph{et al.} presents a new network design paradigm that combines the advantages of manual design and neural architecture search (NAS). 
These cumbersome large-scale models demand heavy computation power and fail to be applied on devices with limited resources.

\begin{figure*}[t]
\begin{center}
\includegraphics[width=16cm]{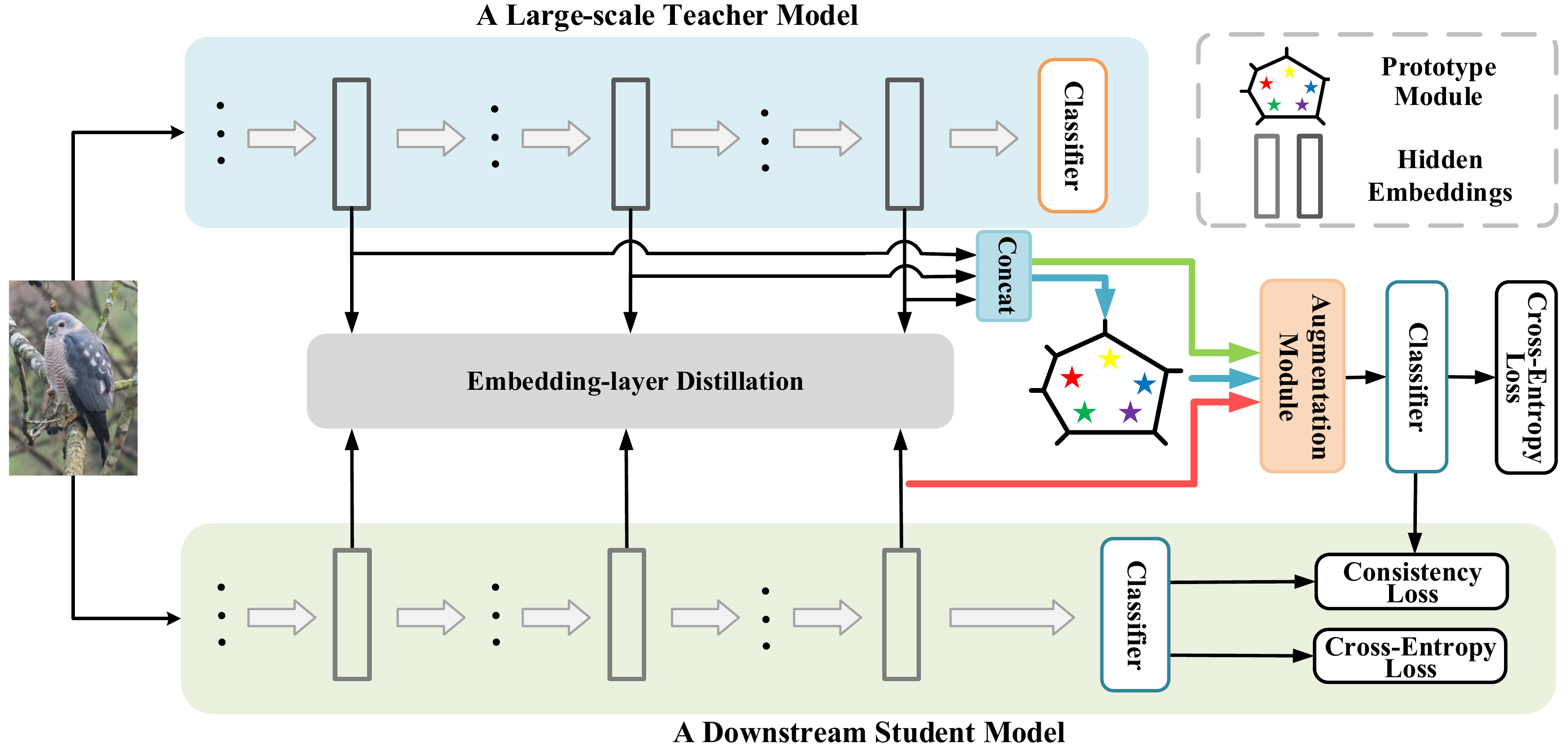}
\end{center}
\caption{Illustration of our proposed prototype-guided cross-task knowledge distillation framework. ProC-KD includes an embedding layer distillation module, a prototype learning module, and a feature augmentation module. The blue arrow in the framework represents generalized representation learning based on prototypes. The green and red arrows indicate that the prototypes are used to enhance the features extracted from the teacher model and the student model, respectively. The augmentation module and the student model share the same classifier.
   }
\label{fig_main_v1}
\end{figure*}

\subsection{Knowledge Distillation}
Knowledge distillation is a model compression technology that transfers the knowledge from a larger deep neural network into a small network. 

The methods of knowledge distillation are mainly divided into response-based knowledge distillation \cite{hinton2015distilling,muller2019when}, feature-based knowledge distillation \cite{romero2015fitnets,huang2017like}, and relation-based knowledge distillation \cite{yim2017fps,park2019relational}. 

The main idea of response-based knowledge distillation is to directly transfer the last output layer neural response of the teacher model. Hinton \emph{et al.} \cite{hinton2015distilling} and Ba \emph{et al.} \cite{ba2014do} propose to shift the knowledge by learning the probabilities distribution via softened labels. However, this method depends on the class probability distribution. The effective method to this is to distill the feature-based or the relation-based knowledge from the teacher model. 
The goal of feature-based knowledge distillation is to match the intermediate representation of the student model with the teacher model. Fitnets \cite{romero2015fitnets} initially introduce intermediate representations learning, in which hints are defined as the outputs of a teacher’s hidden layer to improve the student’s learning process. Inspired by \cite{romero2015fitnets}, a variety of feature-based knowledge distillation methods \cite{zagoruyko2016paying,passalis2018learning,chen2021cross} are proposed. 

Relation-based knowledge distillation methods explore the relationships between different layers \cite{yim2017fps} or data samples \cite{park2019relational}.

KD has also been extensively studied in Transformer-based language models \cite{sanh2019distilbert,sun2019patient,jiao2020tinybert}. While previous knowledge distillation methods for Transformer-based models mainly focus on the NLP domain and the task of the teacher model is the same as the student model. Ye \emph{et al.} \cite{ye2020distilling} deal with a scenario distilling the knowledge from a cross-task teacher. However, it lags in the representation of the invariant intrinsic object and is a two-stage distillation method

Different from existing works, in this paper, we explore the scenario of learning the intrinsic local-level features and reusing the knowledge of large-scale models for different downstream tasks in a cross-task manner.


\section{Main Approach}

Our approach aims to distill the knowledge in the large-scale model to different downstream small models. The label space of the student model is different from the teacher model, which is called a cross-task knowledge distillation. 
Existing same-task knowledge distillation methods mainly directly mimic the final prediction or the intermediate layers of the teacher model, which transfer the global features and are task-specific. The local intrinsic representations can greatly benefit the cross-task student model in understanding the novel dataset of the downstream task.
To improve the generalization ability of the downstream model, we propose a prototype-guided cross-task knowledge distillation method to transfer the invariant intrinsic knowledge from large-scale teacher to small student model as shown in Fig.~\ref{fig_main_v1}. The key module of our method contains a prototype learning module and a feature augmentation module.


\begin{figure*}[t]
\begin{center}
\includegraphics[width=18cm]{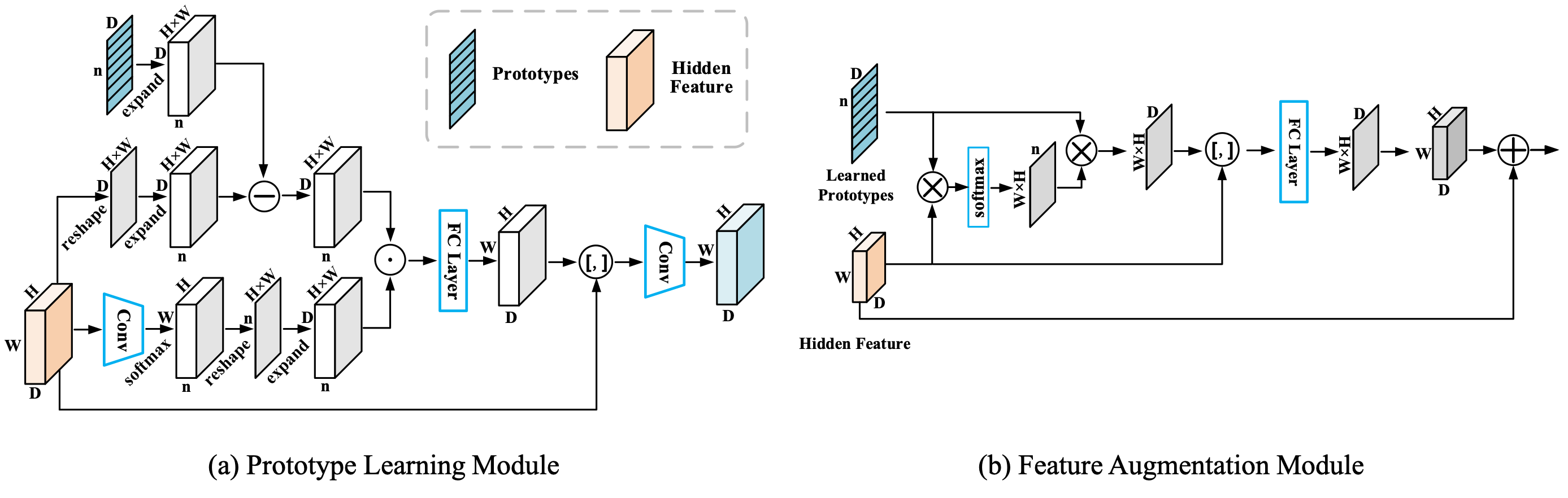}
\end{center}
   \caption{The schematic illustration of (a) prototype learning module and (b) feature augmentation module. Conv and FC Layer separately indicates convolution and fully-connected layer. $\ominus$, $\odot$, $\otimes$, $\oplus$, and [,] denote the residual operation, element-wise multiplication, matrix multiplication, element-wise addition, and concatenation operation, respectively.}
\label{fig_prototype_aug}
\end{figure*}

\subsection{Prototype-based Representation Learning}
Compared to previous methods that directly mimic the final predicted logits or the intermediate layers of the teacher model, our approach designed a module to learn the intrinsic representation. 
Recent studies \cite{snell2017prototypical,liu2019prototype} have demonstrated that constructing prototype learning in models can help to solve novel dataset problems. 
The category-specific information can be captured by prototype learning. Inspired by this idea, we propose a prototype-guided module for the teacher-student distillation architectures to learn the generalized representations with the guidance of prototypes.

The architecture of the prototype learning module is shown in Fig.~\ref{fig_prototype_aug} (a). The forward process is first to align the prototypes with the input feature, then reconstruct the prototype-related feature with the attention mechanism, and finally, aggregate the reconstructed attention features with the input features. The whole process can be divided into three sub-processes, which are \emph{Alignment}, \emph{Attention}, and \emph{Aggregation}.

Specifically, before feeding forward the two-dimensional (2D) hidden layer feature extracted by the Transformer-based model, we reshape it to $F\in \mathbb{R} ^{D \times H \times W} $, where $D$, $H$, and $W$ represent the feature dimension, height, and width, respectively. 

The prototypes are defined as $P$ ( $p_i \in \mathbb{R}^D, i=1,2,...,n $).
In the \emph{Alignment} sub-process, both the defined prototype tensor matrix and the input feature tensor matrix are expanded to $n \times D \times (W \times H) $, and we align them with residual operation $(F-P)$. In the \emph{Attention} sub-process, we calculate the feature descriptors $V$ based on the attention maps, which can be expressed as follows:
\begin{equation}
  V_{i} = \sum_{j=1}^{WH}   \frac{e^{L_{ji}}}{\sum_{i=1}^{n}e^{L_{ji}}}\left ( F_{j} - p_{i}\right ).
\label{equation_v}
\end{equation}

In the \emph{Aggregation} sub-process, we first concatenate the feature descriptors $V$ and the input features, then transform the result with a nonlinear transformation block $f$. It can be formulated as follows:
\begin{equation}
  O_{pro} = f\left ( concat\left [ F, \ \ V_{r}W_{p}+b_{p} \right ] \right ),
\label{equation_proto}
\end{equation} 
where $V_r$ is the reshaped feature descriptors $V$, $W_p$ and $b_p$ are the weight and bias of the fully-connected layer, respectively, and $concat[,]$ indicates the concatenation operation. The shape of the output feature $O_{pro}$ is the same as the input feature $F$.

Through the above
processes. The generalized representation of the prototypes could be learned from the input feature of the large-scale teacher model. Upon such generalized prototypes, we seek to enhance the student features with these prototypes.

\subsection{Feature Augmentation with Prototypes}

To improve the generalization ability of student models of different downstream tasks. The learned generalized prototypes 
are used to enhance the feature in the feature augmentation module, as shown in the right part of Fig.~\ref{fig_main_v1}. The main idea is to enrich the feature which is more related to the prototype representation while suppressing the unrelated feature. 

Fig.~\ref{fig_prototype_aug} (b) shows the architecture of feature augmentation module. 
The forward process is first to pay attention to the feature related to the prototype representation, then enhance the input feature with the prototype-related feature. 
The whole feature augmentation process can also be broken into two sub-processes, which are \emph{Attention} and \emph{Augmentation}. 

Concretely, for the learned prototypes  $P_l \in \mathbb{R}^{n \times D}$ and the hidden features $F_h \in \mathbb{R}^{t \times D}$. In the \emph{Attention} sub-process, we first encode the prototypes and input features respectively. Attention map $A$ is obtained by calculating the softmax of cross-product between learned prototypes $P_l$ and encoded feature $F_e$, which can be expressed as follows:
\begin{equation}
  A = softmax\left (F_{e} P_{l}^\mathsf{T} \right),
\label{equation_softmax}
\end{equation}
and then the attention feature is obtained by calculating the cross-product between the attention map and the prototypes. 

In the \emph{Augmentation} sub-process, we concatenate the attention feature with the encoded input features. The fully-connected layer is applied to transform the result. The original input hidden layer feature is enhanced with the prototype-related feature through element-wise sum operation. The whole feature augmentation process can be written as:

\begin{equation}
  O_{aug} =ReLU(\Phi \left ( concat\left [ F_{e}, \ \ AP_{l} \right ] \right )+F_{r}),
\label{equation_aug}
\end{equation}
where $O_{aug}$ is the output of the feature augmentation module,  $\Phi$ indicates the function of the fully-connected layer, $F_r$ is the reshaped of $F_h$, $W_h$ is the linear transform weight matrix, and $concat[,]$ indicates the concatenation operation. 
At last, the enhanced feature is input to the classifier shared with the student model to predict the categories.

\subsection{Cross-task Knowledge Distillation}

In this paper, the knowledge distillation of student models and teacher models in different label spaces is defined as cross-task knowledge distillation. we proposed the prototype learning module and the feature augmentation module to guide the training of the student model in cross-task distillation scenarios and improve its generalization ability. Above we have described the designed prototype module and feature enhancement module respectively.
Here, we also design some loss functions to constrain the training of cross-task knowledge distillation.
The distillation training loss function includes the embedding-layer knowledge distillation loss function and the prototype learning loss function.

For the embedding-layer knowledge distillation, followed by \cite{jiao2020tinybert}, we distill both the knowledge of attention maps and hidden state features from the large-scale Transformer-based teacher model. Assuming that we are distilling the knowledge from a $m$ layers Transformer-based teacher model to the $n$ layers Transformer-based student model. We need to select $n$ out of $m$ layers from the teacher model. Specifically, for the attention map, the student learns to fit the selected multi-head attention maps in the teacher network, and the loss function for attention-based distillation can be defined as follows:
\begin{equation}
	L_{emb} = \sum_{j=1}^{n}\sum_{i=1}^{h}MSE(A_{i}^{S}, \ A_{i}^{T}) + \sum_{j=1}^{n}MSE(F_{i}^{S}W_{h}, \ F_{i}^{T}),
\label{equation_attn_loss}
\end{equation}
where $T$ indicates the teacher model, $S$ refers to the student model,  $h$ is the attention head number, $A_i \in \mathbb{R}^{l \times l}$ means the attention matrix corresponding to the $i$-th head of teacher or student, $l$ is the input token length, $F^S \in \mathbb{R}^{l \times d'}$ is the student hidden feature and 
$F^T \in \mathbb{R}^{l \times d}$ is the hidden feature of the teacher model. $d$ and $d'$ denote the hidden embedding sizes of the teacher model and student model, respectively. $W_h \in \mathbb{R}^{d' \times d}$ is a learnable transformation weight matrix, which transforms the hidden layer features of the student model into the same dimensions as the features of the teacher model, and $MSE(\cdot)$ indicates the mean squared error loss function.

For the prototype learning, we defined the consistency loss function $ L_{con}$ and the classification loss function $ L_{procls}$. 
The consistency loss is obtained by calculating the Kullback-Leibler (KL) divergence loss between the predicted logits $ y_{con}$ from the prototype augmentation module and the predicted logits $ y_{stu}$ from the student model, $ L_{con} = \mathcal{H}(y_{con}, y_{stu})$).
The classification loss is the softmax cross-entropy loss between prototype prediction $ y_{con}$ and ground-truth labels $y$. Thus the loss function of the prototype learning can be expressed as:

\begin{equation}
\begin{split}
  L_{pro}&=L_{con}+L_{procls},
  \end{split}
\label{proto_loss}  
\end{equation}
 
In addition to the embedding layer feature distillation loss function and the prototype learning loss function, we define the student model loss function as $ L_{stu}$.
The joint training loss function for our cross-task knowledge distillation can be expressed as:
\begin{equation}
\begin{split}
  L_{total}&=\lambda_{emb}L_{emb}+\lambda_{pro} L_{pro}+\lambda_{stu}L_{stu},
  \end{split}
\label{trans_total_loss}  
\end{equation}
where $ \lambda_{emb}$, $ \lambda_{pro} $, and $ \lambda_{stu} $ are the weights of the embedding layer feature distillation loss, the prototype learning loss, and the student model loss, respectively. 

\section{Experiments}
To evaluate the general effectiveness of our method of distilling knowledge from the large-scale models in the cross-task scenarios, we conducted experiments on cross-task knowledge distillation and standard same-task knowledge distillation settings. Experiments were carried out on different visual tasks, \emph{e.g.}, image classification and object detection for each setting. In this paper, the knowledge distillation scheme is set as offline distillation, which means the weights of the teacher network are frozen during the training process.

\subsection{Cross-Task Knowledge Distillation}

\begin{table*}[t]
\begin{center}
\caption{The mean accuracy (\%) of three cross-task image classification knowledge distillation. the cross-domain image classification task is performed on the Office-Home dataset. The LT-CIFAR indicates long-tailed CIFAR 100. FBKD-ProC-KD is our method by plugging the proposed ProC-KD into the FBKD method. The teacher models are all trained on ImageNet-1K \cite{deng2009imagenet} and fixed the weight during distillation training.}
\label{img_cls_result}
\begin{tabular}{l|l|l|c|cc|ccc}
\hline
\multirow{2}{*}{Teacher (param.)} & \multirow{2}{*}{Method} & \multirow{2}{*}{param.} & Standard
  & \multicolumn{2}{c|}{Long-tailed} & \multicolumn{3}{c}{Cross-domain} \\ \cline{4-9}   
                                                                               &     &    & CIFAR-100 & LT-CIFAR&iNaturalist 2018 & Rw$\rightarrow$Ar& Rw$\rightarrow$Cl & Rw$\rightarrow$Pr       \\ \hline
\multirow{4}{*}{ViT-B\cite{dosovitskiy2020image} (86M)} & Student Model   &  43M   &78.84   &  55.83      & -         &20.85         &16.91        &34.85        \\       
															   & RKD \cite{park2019relational}   &  43M   & 87.13  &  76.52      &      -      & 57.31        & 40.64       & 73.50       \\
                                                               & FBKD \cite{jiao2020tinybert}              &  43M   &86.16           &  72.69         & 57.14    & 60.28  &  40.02  &75.51        \\
                                                               & FBKD-ProC-KD (Ours)  &  43M   & \textbf{87.46} & \textbf{78.32} & \textbf{58.67}& \textbf{61.41}&\textbf{40.96}&\textbf{76.83}  \\ \hline
\multirow{4}{*}{Swin-L\cite{radosavovic2020designing} (197M )} & Student Model    &  110M  &78.90    &  41.48              & -          & 26.87 &20.51              & 43.32        \\
															   & RKD \cite{park2019relational}   &  110M   &83.99   &58.94       & -
															             &22.95         & 15.50       & 30.71      \\
                                                               & FBKD \cite{jiao2020tinybert}               &  110M  &83.63           &   57.83        & 70.19         &41.29          & 29.03        & 63.40        \\
                                                               & FBKD-ProC-KD (Ours)  &  110M  & \textbf{84.21} & \textbf{68.23} & \textbf{72.11}& \textbf{42.16}&\textbf{30.24}&\textbf{64.27}  \\ \hline
\end{tabular}
\end{center}

\end{table*}

\begin{figure*}[!h]
\begin{center}
\includegraphics[width=18cm]{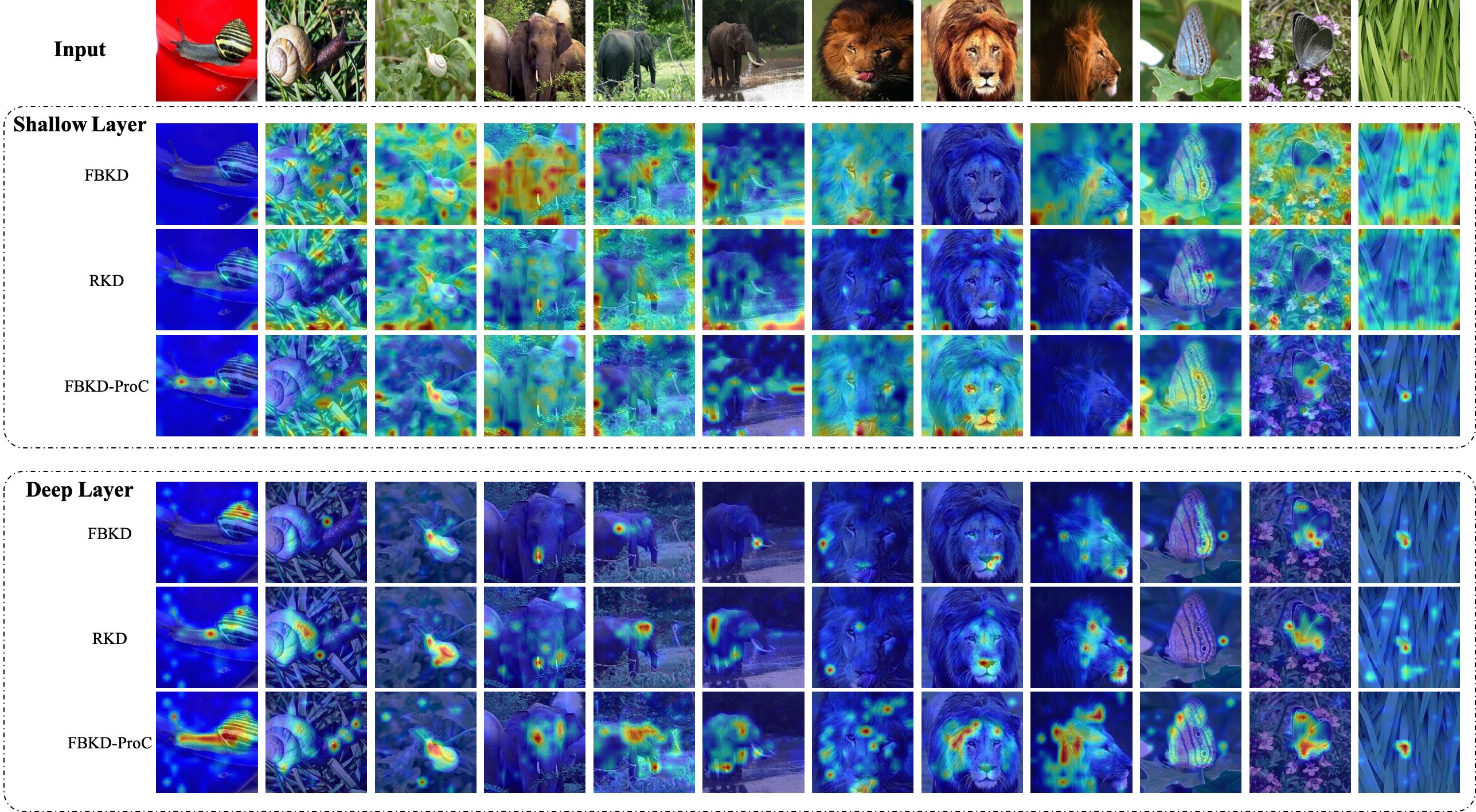}
\end{center}
   \caption{The comparison of attention maps in FBKD and FBKD-ProC-KD (ours) by using the Transformer Interpretability method \cite{chefer2021transformer}. Here, the second layer and the last layer of the ViT are selected as the shallow layer and the deep layer, respectively. }
\label{attention_map}
\end{figure*}

\subsubsection{Image Classification}
We carried out experiments on the Transformer-based model in three downstream tasks, including standard image classification, long-tailed image classification, and cross-domain image classification. Here, all the teacher models are trained on ImageNet-1K \cite{deng2009imagenet} dataset in our experiments. 

\textbf{Datasets.}
CIFAR-100 \cite{rebuffi2017learning} consists of 50,000 training images and 10,000 validation images. It contains 100 categories and each class contains 600 images. The size of each image is 32 $\times$ 32.
Following \cite{cao2019learning,cui2019class}, the long-tailed CIFAR-100 is created by reducing the number of training samples for each class, but with the verification set unchanged. We define an imbalance ratio $\beta$, \emph{i.e.}, $\beta = N_{max}/N_{min}$. $\beta$ represents the ratio of sample sizes between the most frequent class and the least frequent class. In this way, sample sizes decay exponentially between classes. In our experiment, we set the imbalance ratio to 10.
iNaturalist 2018 \cite{van2018inaturalist} contains over 450,000 training images from 8,142 different species of birds, mammals, reptiles, and plants among others. Compared with ImageNet and other image classification datasets, iNaturalist exhibits a long-tail distribution, and many species have relatively few images. We used the official training and validation split in the experiment, with 437,513 images for training and 24,424 images for validation.
The Office-Home \cite{venkateswara2017deep} dataset has been created to evaluate domain adaptation methods for image classification. It consists of 15,500 images from four different domains: Artistic images (Ar), Clip Art (Cl), Product images (Pr), and Real-World images (Rw). Each domain in this dataset contains 65 categories, and the images are from office or home scenes. In our experiments, the Real-World images (Rw) are used as the training set, and the other domains are used as the test sets.

\textbf{Implementation Details.}
We conduct our experiments on the well-known ViT \cite{dosovitskiy2020image} and Swin-Transformer \cite{liu2021swin} models. 

For the ViT, the teacher model is a 12-layer ViT-B model and the student model is a 6-layer small ViT model. The indexes of hidden layers selected for distillation in the teacher model are [2, 4, 6, 8, 10, 12]. Both the attention map knowledge and hidden layer feature knowledge are distilled. All the coefficients of the loss function Eq. \ref{trans_total_loss} are 1 except for the $ \lambda_{emb}$ is 0.3. In the training phase, the $4$-th, $8$-th, and $12$-th layer hidden features are selected to concatenate and then input to the prototype learning module and augment module to train the prototypes. The number of prototypes is set to 72. The AdamW optimizer is used with a learning rate of 5e-4 and a weight decay of 0.05. The input size of the image is $224\times 224$ and the batch size is set to 32 for each GPU.

For the Swin-Transformer, The teacher model is 24-layer Swin-L, we distill the knowledge from the middle 18-layers to 6-layers, and the student model is a 12-layers Swin-Transformer. The last two hidden layers features of the backbone are selected for the prototype learning module and augmentation module after the concatenation operation. 
We use the AdamW optimizer with an initial learning rate of 5e-4 and a weight decay of 0.05. The training batch size is set to 64 for each GPU.

All the experiments are run on 8$\times$ Nvidia Tesla V100 GPUs (32GB VRAM, PCIe connection). We use NCCL for multi-node parallel training. Gradient accumulation is also applied to reduce multi-GPU communication overheads.

\begin{table*}[t]
\begin{center}
\caption{Results (\%) of cross-task object detection knowledge distillation on Cityscapes and FoggyCityscapes. The teacher models are all trained on the COCO \cite{lin2014microsoft} dataset and fixed the weight during distillation training.}
\label{table:fogcity_results}
\begin{tabular}{l|cccccccc|c}
\hline
\multicolumn{10}{c}{\textbf{Cityscapes}}    
\\
\hline
Method        & person     & rider         & car         &truck & bus      & train       & motorcycle       & bicycle       & mAP \\ \hline
Student Model & 53.1        & 55.1        & 70.1        & 31.3 &56.1     & 31.6        & 40.2        &44.6         &47.8   \\
CWD \cite{shu2021channel} &   63.2          & \textbf{65.2}           &  77.7           & 48.6         & 72.8        &  49.8    &\textbf{54.2}   & 58.1     & 61.2 \\
ProC-KD (Ours)       &\textbf{63.9}&65.1&\textbf{77.8}& \textbf{51.9}&\textbf{74.3}&\textbf{51.7}&52.5&\textbf{59.9}&\textbf{62.1} \\ \hline
\hline
\multicolumn{10}{c}{\textbf{FoggyCityscapes}}    
\\ \hline
Student Model & 40.8        & 40.3        & 63.0        & 27.8      & 42.3        & 11.6        &27.3         & 31.8        &35.6  \\
FBKD \cite{jiao2020tinybert}          & 52.0        &54.1         & 68.6        &37.5       &53.9         & 35.7        & 40.6        &49.0         &48.9  \\
CWD \cite{shu2021channel}          & 53.3        &55.8         & 71.9        &37.5       &57.1         &46.8         &43.2         &50.9         &52.1  \\
FBOD \cite{zhang2020improve}     & 52.1            &51.2             & 69.3            & 36.0          & 53.7            & 42.6            &41.4             & 45.4            & 48.9 \\
\hline
\hline

FBKD-ProC-KD (Ours)     &   51.8          &  55.0           &  68.8           & 38.7         & 53.4        &  47.1       & 38.8     & 45.2            &49.9  \\
ProC-KD (Ours)       &\textbf{53.8}&\textbf{57.9}&\textbf{73.1}& \textbf{40.3}&\textbf{57.7}&\textbf{51.2}&\textbf{44.2}&\textbf{51.5}&\textbf{53.7} \\ \hline
\end{tabular}
\end{center}
\end{table*}

\textbf{Results and Analysis.}
Table~\ref{img_cls_result} shows the experimental results on the standard, long-tailed, and cross-domain image classification tasks. We compare it with Relation Knowledge Distillation (RKD) \cite{park2019relational} by reimplementing it in our experimental setting.
Followed by Tinybert \cite{jiao2020tinybert}, FBKD is a feature-based knowledge distillation method for the Transformer-based model that distills the knowledge from embedding layers and attention maps to the student model. FBKD-ProC-KD is our method by plugging the proposed ProC-KD into the FBKD method.

{\it Standard Image Classification.} Compared with the baseline method, our ProC-KD improves by 1.3\% and 0.6\% on ViT and Swin-Transformer, respectively. 
This demonstrates that our ProC-KD method can promote the prototypes to learn the generalized representation and improve the performance of the student model in cross-task image classification scenes. 

{\it Long-tailed Image Classification.} For the long-tailed CIFAR-100 dataset, our ProC-KD improves the performance by 5.6\% and 10.4\% on ViT and Swin-Transformer, respectively. For the iNaturalist 2018 dataset, our ProC-KD separately improves the performance by 1.5\% and 1.9\% on ViT and Swin-Transformer. This demonstrates that ProC-KD can improve the generalization ability of the student model in the long-tailed image classification tasks. 

{\it Cross-domain Image Classification.} The cross-domain image classification experiment is conducted on the Office-Home dataset. As can be seen, compared with the FBKD baseline, ProC-KD improves the performance by 1.3\% on domain shift Rw$\rightarrow$Pr and by 1.2\% on hardest domain shift Rw$\rightarrow$Cl. This demonstrates that distilling the knowledge from the model trained on large-scale datasets can improve the performance of the student model in the cross-domain scene, and our ProC-KD can further improve the generalization ability.

Fig. \ref{attention_map} shows the visualization results of the attention maps in FBKD and our FBKD-ProC-KD. As can be seen, compared with the FBKD baseline, the attention map of our method FBKD-ProC-KD focuses more on objects in both the shallow layer and deep layer. It indicates that our ProC-KD method can promote the training of the student model in both the shallow layers and the deep layers.

\begin{figure*}[!h]
\begin{center}
\includegraphics[width=18cm]{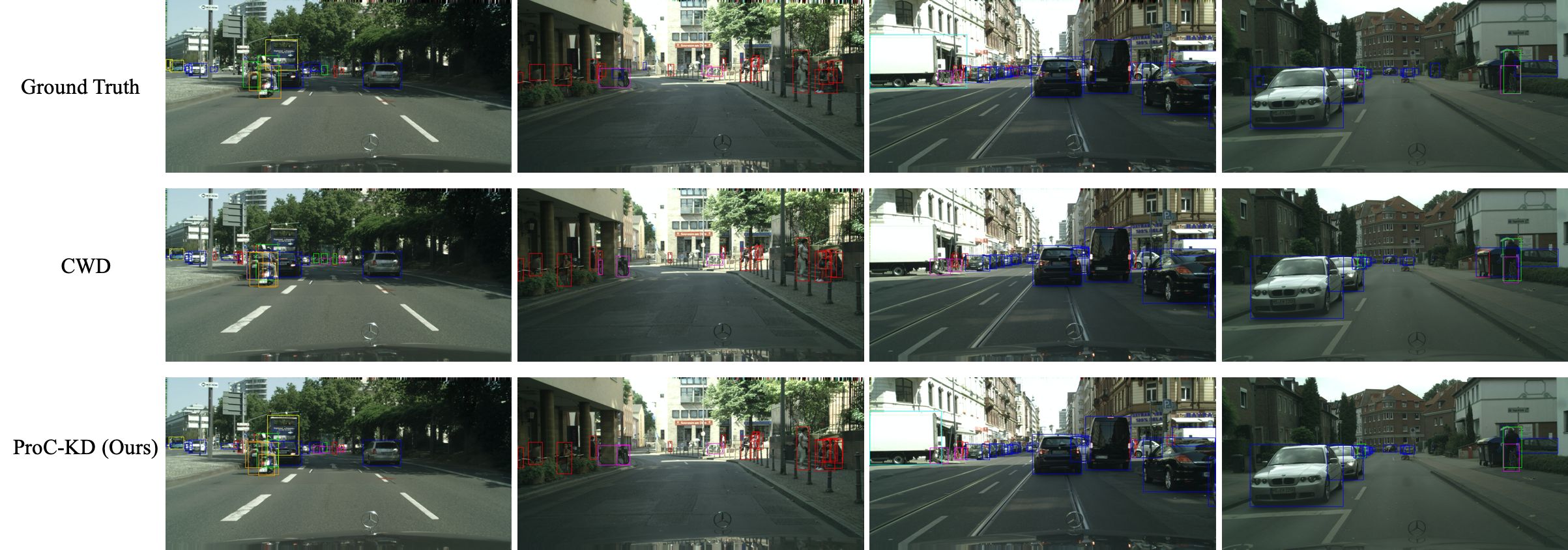}
\end{center}
   \caption{Qualitative results on Cityscapes. Compared with CWD
   baseline, our ProC-KD method could localize and recognize objects accurately, e.g., the {\color[rgb]{1.00,1.00,0.00} bus}, {\color[rgb]{1.00,0.00,1.00} bicycle}, {\color[rgb]{0.00,1.00,1.00} truck}, {\color[rgb]{1.00,0.00,0.00} person}. }
   
\label{cityscpae_visual}
\end{figure*}

\begin{figure*}[!h]
\begin{center}
\includegraphics[width=18cm]{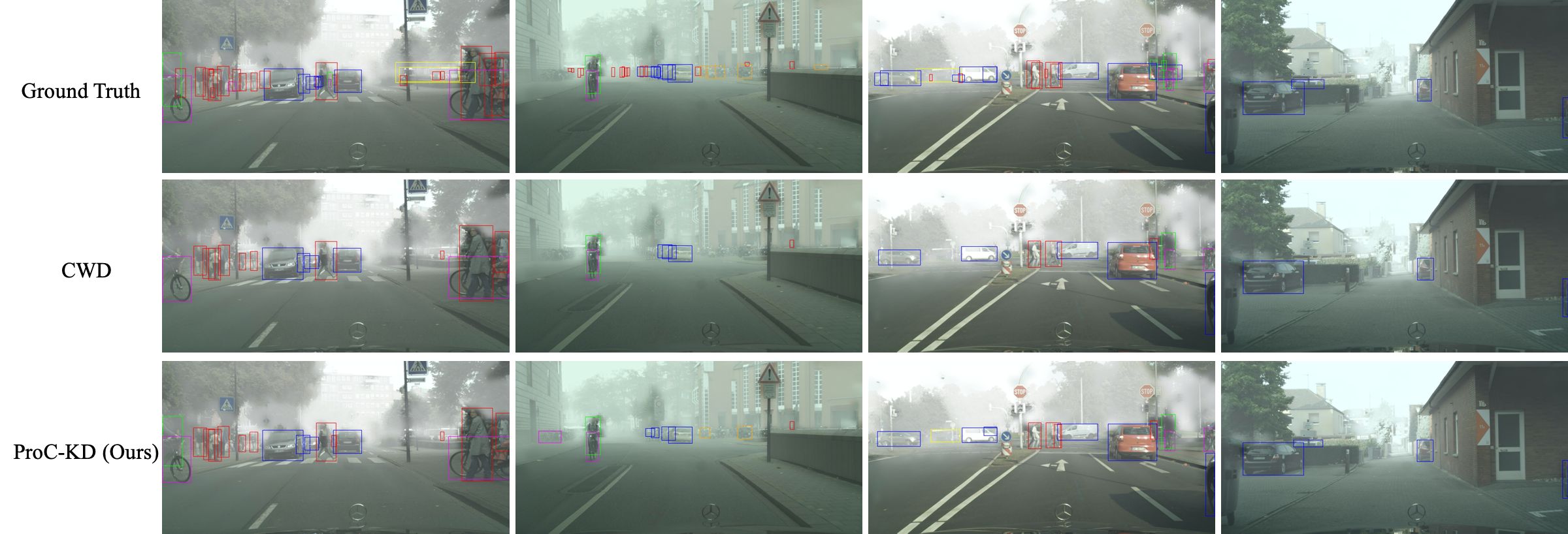}
\end{center}
   \caption{Qualitative results on FoggyCityscapes. 
   Compared with CWD \cite{shu2021channel} baseline, our ProC-KD method could localize and recognize objects accurately in foggy images, e.g., the {\color[rgb]{0.00,1.00,0.00} rider}, {\color[rgb]{1.00,0.00,1.00} bicycle}, {\color[rgb]{1.00,1.00,0.00} bus}, {\color[rgb]{0.00,0.00,1.00} car}. }
\label{foggycity_visual}
\end{figure*}

\subsubsection{Object Detection}
We also conduct experiments on standard object detection and cross-domain object detection. For the cross-domain object detection task, we only take the source domain as the training set and the target domain as the testing set. Here, all the teacher models are trained on COCO \cite{lin2014microsoft} dataset and frozen the weight during the distillation training. 

\begin{table*}[!h]
\begin{center}
\caption{Results (\%) of cross-task object detection knowledge distillation on cross-domain of Cityscapes $\rightarrow$ FoggyCityscapes. Here, We train the model on the training data of Cityscapes and test the model on the test dataset of FoggyCityscapes. The teacher models are all trained on the COCO \cite{lin2014microsoft} dataset and fixed the weight during distillation training.} 
\label{da_fogcity_results}
\begin{tabular}{l|cccccccc|c}
\hline
Method        & person     & rider         & car         &truck & bus      & train       & motorcycle       & bicycle       & mAP \\ \hline
Student Model & 22.3        & 17.6        & 18.0        & 4.5      & 9.1        & 0.0        &9.1         & 20.5        &12.6  \\
CWD\cite{shu2021channel}   & 41.0        &48.7         & 51.3        &20.2       &33.1         & 10.6        & 30.8        &44.4         &35.0  \\
ProC-KD (Ours)       &\textbf{41.3}&\textbf{48.9}&\textbf{51.7}& \textbf{24.6}      &\textbf{33.8}&\textbf{13.5}&\textbf{32.6}&\textbf{44.4}&\textbf{36.3} \\ \hline
\end{tabular}
\end{center}
\end{table*}

\begin{table*}[!h]
\begin{center}
\caption{Results (\%) of cross-task object detection knowledge distillation on cross-domain of Daytime-sunny $\rightarrow$ Night-rainy and  Daytime-sunny $\rightarrow$ Dusk-rainy. Here, We train the model on the Daytime-sunny data and test the model on the Night-rainy and Dusk-rainy data. The teacher models are all trained on the COCO \cite{lin2014microsoft} dataset and fixed the weight during distillation training.}
\label{rainy_result}
\begin{tabular}{lcccccccc}
\hline
\multicolumn{9}{c}{\textbf{Daytime-sunny$\rightarrow$Night-rainy}}                                                                                                                                                                                                \\ \hline
\multicolumn{1}{l|}{Method}        & bicycle              & bus                  & car                  & motorcycle           & person               & rider                & \multicolumn{1}{c|}{truck} & mAP                  \\ \hline
\multicolumn{1}{l|}{Student Model} & 24.3        & 9.1         & 33.8        & 1.1       & 12.3        & 9.1         &\multicolumn{1}{c|}{ 16.1 }       & 15.1 \\                
\multicolumn{1}{l|}{CWD \cite{shu2021channel}}           & 38.6  &17.1    & 49.4  &\textbf{9.7}  &24.4   &15.6  &\multicolumn{1}{c|}{34.4 }         &27.0  \\                 
\multicolumn{1}{l|}{ProC-KD (Ours)}       &\textbf{40.9}&\textbf{18.3}&\textbf{49.4}& 8.6      &\textbf{26.1}&\textbf{18.2} &\multicolumn{1}{c|}{\textbf{35.7}} &\textbf{27.9}                      \\ \hline
\hline
\multicolumn{9}{c}{\textbf{Daytime-sunny$\rightarrow$Dusk-rainy}}                                                                                            \\ \hline
\multicolumn{1}{l|}{Student Model} &40.6         & 14.9         & 66.0        & 11.5       & 25.8        & 15.2         & \multicolumn{1}{c|}{39.7}        & 30.5 \\
\multicolumn{1}{l|}{CWD \cite{shu2021channel}}              &49.9         &34.8         & 73.9        &\textbf{24.0}       &43.9         &32.0         &\multicolumn{1}{c|}{54.7}         &44.7  \\
\multicolumn{1}{l|}{ProC-KD (Ours)}              &\textbf{52.6}&\textbf{36.6}&\textbf{73.3}& 21.6      &\textbf{46.5}&\textbf{31.6}&\multicolumn{1}{c|}{\textbf{54.6}}&\textbf{45.2} \\ \hline\end{tabular}
\end{center}
\end{table*}

\begin{figure*}[!h]
\begin{center}
\includegraphics[width=18cm]{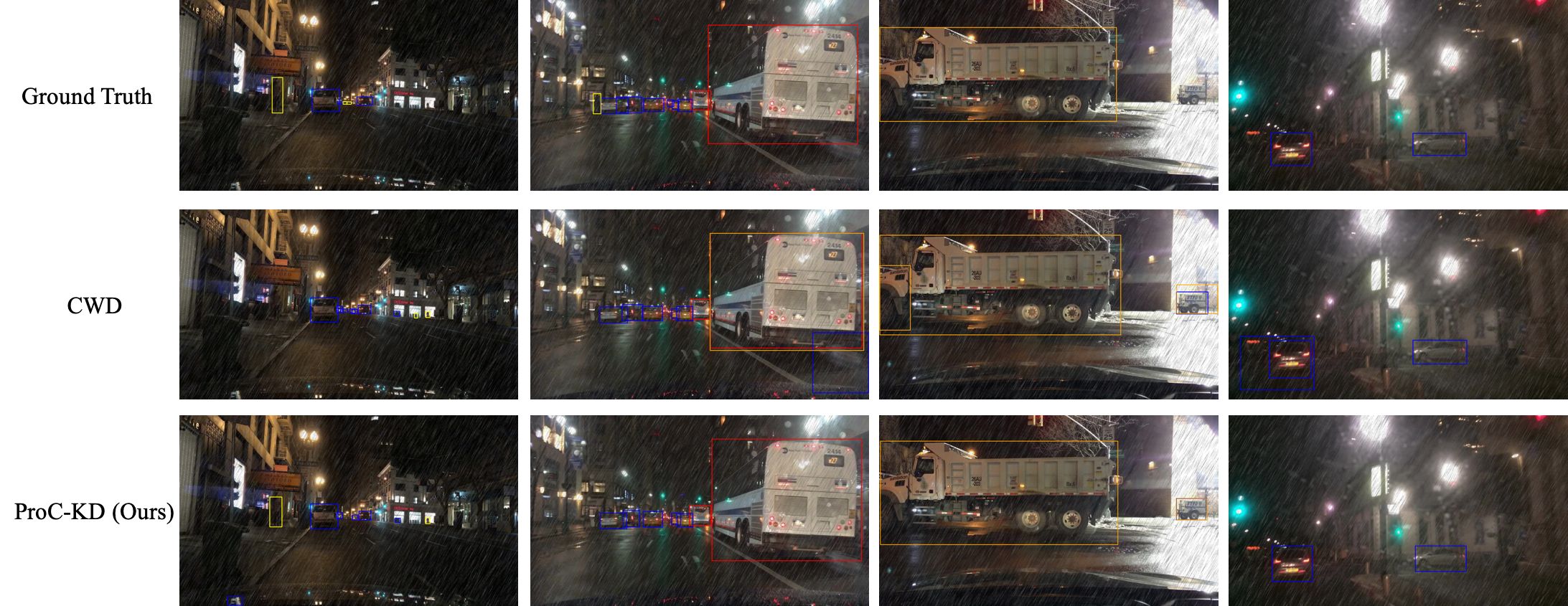}
\end{center}
   \caption{Qualitative results on Daytime-sunny$\rightarrow$Night-rainy. 
   Compared with CWD \cite{shu2021channel} baseline, our ProC-KD method could localize and recognize objects accurately in foggy images, e.g., the {\color[rgb]{1.00,1.00,0.00} person}, {\color[rgb]{1.00,0.00,0.00} bus}, {\color[rgb]{1.00,0.73,0.00} truck}, {\color[rgb]{0.00,0.00,1.00} car}. }
\label{night_rainy_visual}
\end{figure*}

\begin{figure*}[!h]
\begin{center}
\includegraphics[width=18cm]{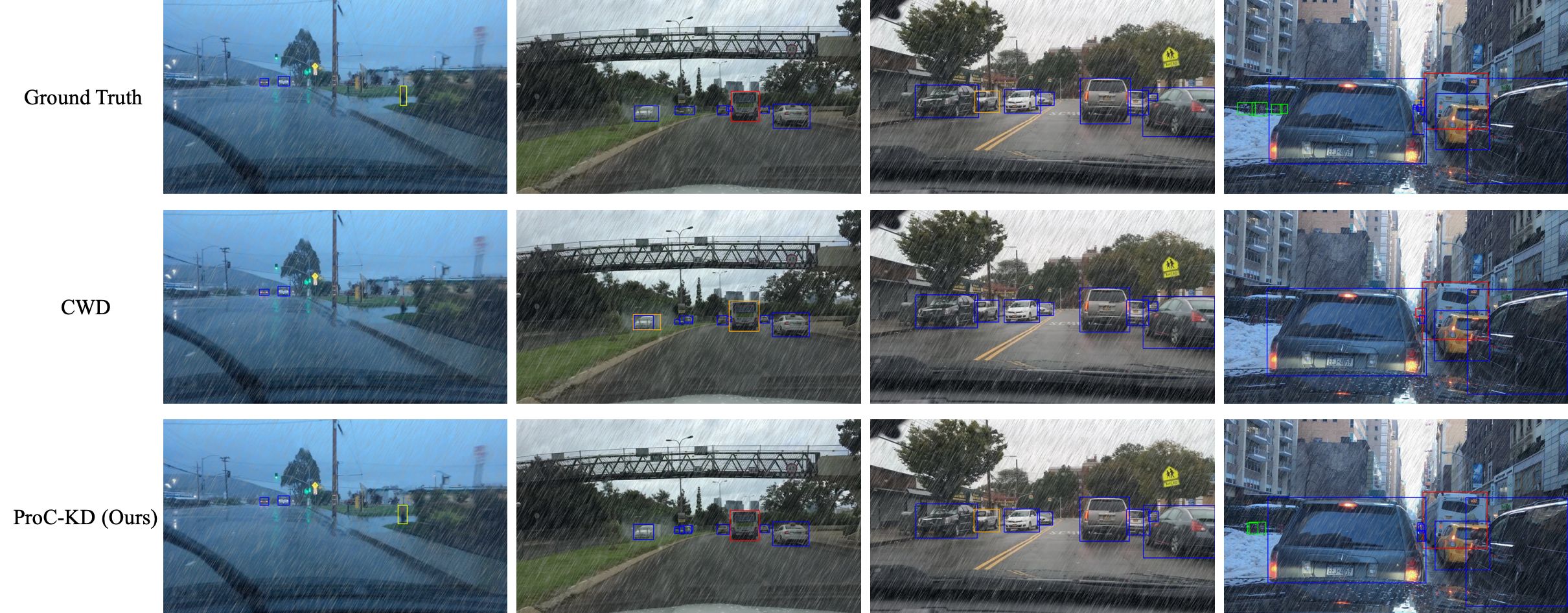}
\end{center}
   \caption{Qualitative results on Daytime-sunny$\rightarrow$Dusk-rainy. 
   Compared with CWD
baseline, our ProC-KD method could localize and recognize objects accurately in rainy images, e.g., the {\color[rgb]{1.00,1.00,0.00} person}, {\color[rgb]{1.00,0.00,0.00} bus}, {\color[rgb]{0.00,0.00,1.00} car}, {\color[rgb]{1.00,0.73,0.00} truck}, {\color[rgb]{0.00,1.00,0.00} bycicle}. }
\label{dusk_rainy_visual}
\end{figure*}

\textbf{Datasets.}
Cityscapes \cite{cordts2016cityscapes} is an urban street dataset of 8 categories of objects. It contains 2975 and 500 images in the training and validation set. FoggyCityscapes \cite{sakaridis2018semantic} is a dataset obtained by synthesizing different degrees of fog on Cityscapes \cite{cordts2016cityscapes}. It contains 2975 training and 500 validation images, respectively. Daytime-sunny, Dusk-rainy, and Night-rainy \cite{sakaridis2018semantic} are three street scene datasets under different weather environments collected from the BDD-100k dataset. In our experiments, we select 27,708 images from Daytime-sunny as the training set, and 2,494 and 3,501 images from Night-rainy and Dusk-rainy as the test set for the two scenarios, respectively.

\textbf{Implementation Details.}
For the cross-task knowledge distillation experiment of standard object detection and cross-domain object detection, the teacher model is the Cascade Mask-RCNN with the backbone of 24 layers Swin-Base \cite{liu2021swin} model trained on the COCO \cite{lin2014microsoft} dataset and the student model is the Cascade Mask-RCNN with the backbone of 12-layers Swin-Tiny model. 
In addition to the distillation learning of the hidden layer knowledge in the teacher model, the fourth layer feature of the FPN is selected as the input to the prototype learning module and augmentation module for generalized representation learning. The coefficients of $ \lambda_{pro} $ and $ \lambda_{emb} $ are both set to 1. We run the SGD optimizer with the initial learning rate of 0.01 and the parameter decay of 0.0001 for 36 epochs. The batch size is set to 2 for each GPU.

To compare with the state-of-the-art knowledge distillation methods of object detection. We also reimplement some representative feature-based knowledge distillation object detection methods based on our experimental setting, \emph{e.g.}, CWD \cite{shu2021channel} and FBOD \cite{zhang2020improve}.

\textbf{Results and Analysis of Standard Object Detection.}
Table~\ref{table:fogcity_results} shows detection results on Cityscapes and FoggyCityscapes. Here, we reimplement the method of CWD \cite{shu2021channel} and FBOD \cite{zhang2020improve} in our experimental setting. FBKD is a knowledge distillation method that follows Tinybert \cite{jiao2020tinybert}, and we only distill the hidden features of the backbone. We can see that our method boosts the performance under the cross-task knowledge distillation scene significantly.
For Cityscapes, compared to the baseline of CWD \cite{shu2021channel}, our method separately improves the performance by 0.9\%.
For FoggyCityscapes, compared to FBKD and CWD \cite{shu2021channel}, our method separately improves the performance by 1.0\% and 1.4\%. This demonstrates that the generalized prototype representation is helpful for the learning of the student model on object detection.

Fig.~\ref{cityscpae_visual} and Fig. \ref{foggycity_visual} show the visualization of the detection results on Cityscape and FoggyCityscapes respectively. The first row is the ground truth of the objects, the second row is the detection results of the baseline method CWD \cite{shu2021channel}, and the third row is the detection results of ProC-KD. We can see that, compared with CWD, our method ProC-KD could localize and recognize objects accurately in the normal scene and foggy images.

\begin{figure*}[!h]
\begin{center}
\includegraphics[width=17cm]{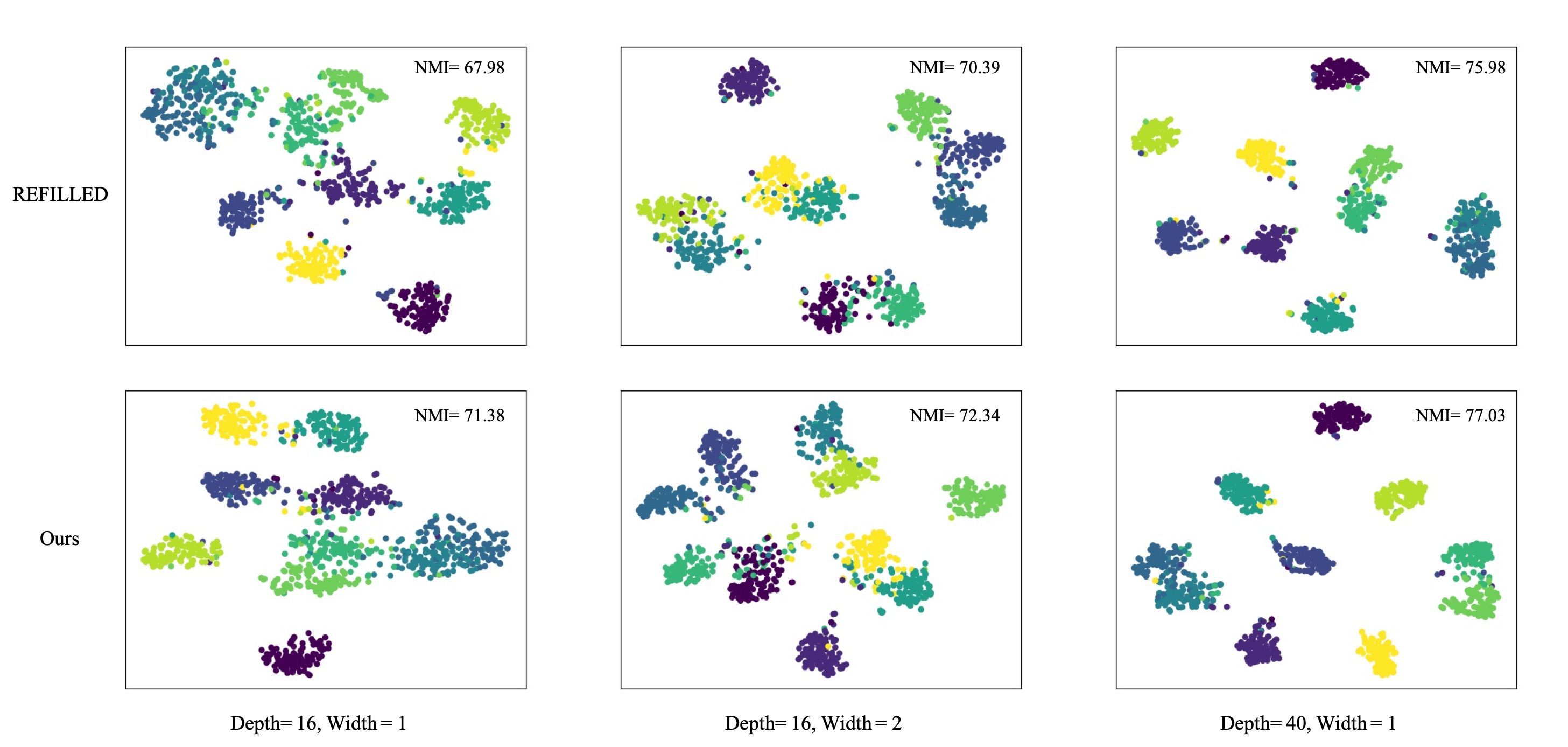}
\end{center}
   \caption{The tSNE of the REFILLED (left) and Ours (right) over 10 classes sampled from CIFAR-10. The larger the NMI value means the better embedding quality.}
\label{stand_tsne}
\end{figure*}

\textbf{Results and Analysis of Cross-domain Object Detection.}
Table~\ref{da_fogcity_results} shows the results on cross-domain object detection of Cityscapes $\rightarrow$ FoggyCityscapes. As we can see, compared with baseline CWD \cite{shu2021channel} our ProC-KD improves the performance significantly by 1.3\%. For most of the object categories, our method outperforms CWD \cite{shu2021channel}. This demonstrated that our ProC-KD method could improve the generalized ability of the student model on cross-domain object detection.

Table \ref{rainy_result} shows the results on cross-domain object detection of Daytime-sunny $\rightarrow$ Night-rainy and Daytime-sunny$\rightarrow$Dusk-rainy. For Daytime-sunny $\rightarrow$ Night-rainy, compared with the CWD \cite{shu2021channel} baseline, our ProC-KD method improves the mAP by 0.9\%. For Daytime-sunny$\rightarrow$Dusk-rainy, our ProC-KD method improves the mAP by 0.5\%. 
The reason that the performance of our ProC-KD is lower than CWD on the object of the motorcycle may be that the number of the ground truth of the motorcycle in the testing set is quite small. It only contains 49 ground truths of motorcycles in the Daytime-sunny $\rightarrow$ Night-rainy testing set and only 110 ground truths in the Daytime-sunny$\rightarrow$Dusk-rainy testing set. 

Fig. \ref{night_rainy_visual} and Fig.~\ref{dusk_rainy_visual} show the visualization of the cross-domain object detection results on the Daytime-sunny$\rightarrow$Night-rainy and Daytime-sunny$\rightarrow$Dusk-rainy respectively. The first row is the ground truth of the objects, the second row is the detection results of the baseline method CWD \cite{shu2021channel}, and the third row is the detection results of ProC-KD. 
Compared with CWD, our ProC-KD could localize and recognize objects in both night rainy images and dusk rainy images accurately.

\subsection{Standard Same-task Knowledge Distillation}
We regard knowledge distillation in which the teacher model and student model share the same label space as standard knowledge distillation. The method we proposed is a general method, which can also be used in the setting of standard knowledge distillation. Here we verify the effectiveness of our method on standard knowledge distillation in the image classification task and object detection task respectively.

\subsubsection{Image Classification} In this part, the teacher model and student model share the same image label space. Here, we also verify the effectiveness of our method on the classification model of CNN structure, we set the teacher model and the student model to be Wide-ResNet which is a CNN-based network structure. By changing the depth and width of the student model, we can get different student models to verify the adaptability of the method to different network structures. The dataset used in the experiment is CIFAR-100 \cite{rebuffi2017learning}. Followed with REFILLED \cite{ye2020distilling}, all teacher models are set as Wide-ResNet with a depth of 40 and width of 2 in these experiments. The accuracy of the teacher model is 74.44\%. Different from the two-stage optimization of REFILLED \cite{ye2020distilling}, our ProC-KD only performs one-stage optimization.

\textbf{Results and Analysis.} Table \ref{stand_cls_dist} shows comparison results between our method and other SOTA distillation methods with different student models. Same with REFILLED \cite{ye2020distilling}, the accuracy of our method is the result of the test on the test set after the training convergence on the training set. The results of other comparison methods are cited from REFILLED \cite{ye2020distilling}. It can be seen from Table \ref{stand_cls_dist} that compared with other knowledge distillation methods, our method achieves the best accuracy in three student models with different structures. 
Compared with REFILLED \cite{ye2020distilling}, our ProC-KD outperforms by 1.3\%, 0.48\%, and 0.5\% in the three student network structures of (depth, width)=(40,1), (depth, width)=(16,2), and (depth, width)=(16,1). This demonstrates the effectiveness of our method in standard image classification knowledge distillation scene.

Figure \ref{stand_tsne} shows the visualization results of the randomly sampled 10 classes embedding features with tSNE\cite{van2012stochastic}. The normalized mutual information (NMI) is used as the criterion to measure the embedding quality, the value is larger means the embedding quality is better. we can see that for the embedding features of 10 categories sampled randomly, our method is more discriminative and has higher NMI values.

\begin{figure*}[!h]
\begin{center}
\includegraphics[width=18cm]{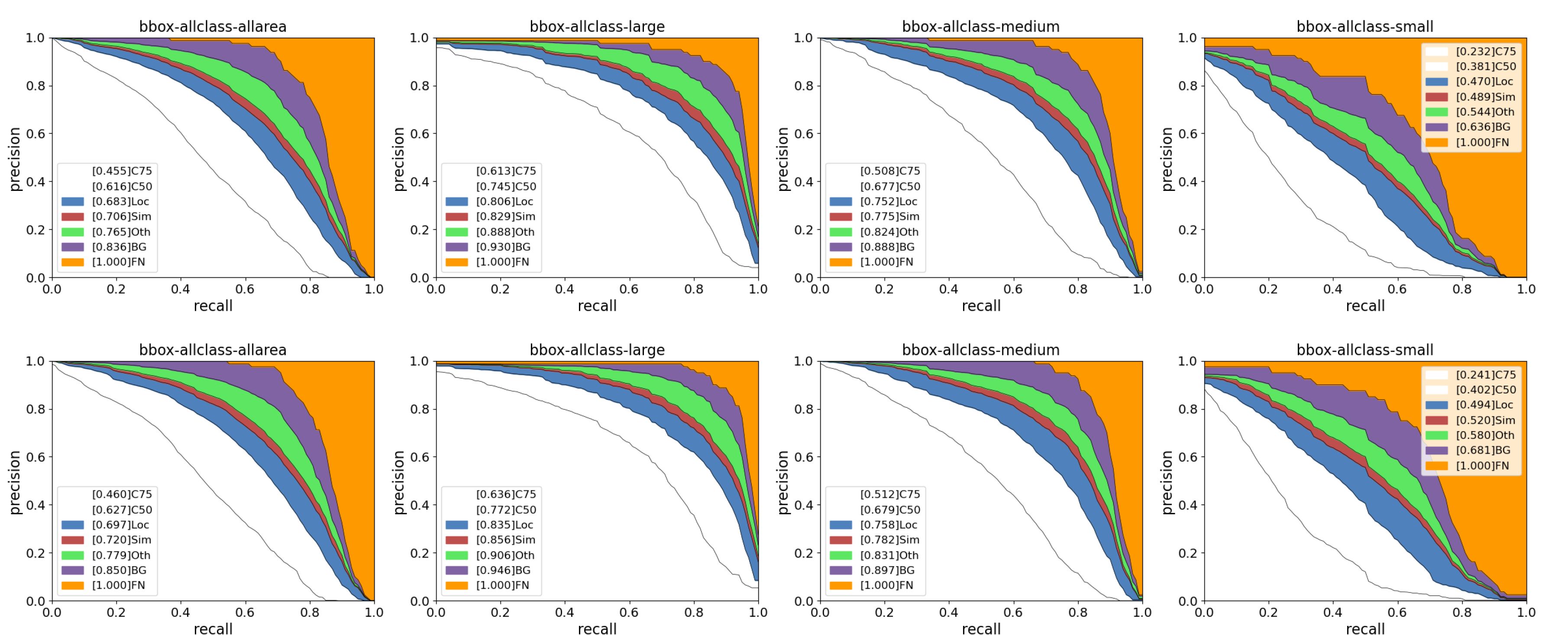}
\end{center}
\caption{The error analysis Precision-Recall curve of all objects, large size objects, medium size objects, and small size objects on the COCO \cite{lin2014microsoft} dataset. The top row is from the baseline CWD \cite{shu2021channel} and the bottom row is from our ProC-KD. Here, C75 indicates the results at 0.75 IoU threshold, C50 indicates the Results at 0.50 IoU threshold, Loc indicates the results after ignoring localization errors, Sim indicates the results after ignoring the similar classes from the same supercategory false positives, Oth indicates the results after ignoring all category confusions, BG indicates the results after ignoring all false positives, and FN indicates the results after ignoring all false negatives.}
\label{std_pr_curve}
\end{figure*}

\begin{table}[t]
\begin{center}
\caption{Results (\%) on the standard image classification knowledge distillation scene. Here, the teacher model and student model share the same CIFAR-100 \cite{rebuffi2017learning} label space.}
\label{stand_cls_dist}
\begin{tabular}{c|ccc}
\hline
Method/(depth, width)       & (40, 1)    & (16, 2) & (16, 1)    \\ \hline
Student      & 68.97        &70.15         & 65.44               \\ \hline
KD \cite{hinton2015distilling}      & 70.46        & 71.87         & 66.54               \\ 
FitNet\cite{romero2015fitnets}  & 68.66        & 70.89         & 65.38               \\
AT \cite{zagoruyko2016paying}      & 69.85        & 71.06       & 65.31                 \\
NST \cite{huang2017like}     & 68.00        & 71.19         & 64.95              \\
VID-I \cite{ahn2019variational}    & 71.51        & 73.31         & 66.32           \\
RKD \cite{park2019relational}      & 72.18        & 72.56       & 65.22             \\
REFILLED \cite{ye2020distilling}      & 72.72        & 74.01        & 67.56            \\ \hline
ProC-KD (Ours)      & \textbf{74.05}        & \textbf{74.49}        & \textbf{68.06}             \\

\hline
\end{tabular}
\end{center}
\end{table}

\begin{table}[t]
\begin{center}
\caption{Results (\%) on the standard object detection knowledge distillation scene. Here, the teacher model and student model share the same COCO \cite{lin2014microsoft} label space. '${\sharp}$' indicates the results that we reimplement with the released code}
\label{stand_obj_det_dist}
\begin{tabular}{c|cccccc}
\hline
Method       & AP             & AP$_{50}$       & AP$_{75}$   & AP$_s$         &AP$_m$            & AP$_l$    \\ \hline
Teacher      & 44.3          & 62.7       & 48.4   & 25.4        & 48.4         & 58.1      \\ 
Student      & 38.4          & 59.0        & 42.0   & 21.5        & 42.1         & 50.3       \\ \hline
Chen \cite{choi2018learning}    & 38.7          & 59.0        & 42.1   & 22.0        & 41.9         & 51.0           \\ 
Wang \cite{wang2019distilling}   & 39.1          & 59.8        & 42.8   & 22.2        & 42.9         & 51.1           \\
Heo \cite{heo2019comprehensive}     & 38.9          & 60.1        & 42.6   & 21.8        & 42.7         & 50.7           \\
FBOD \cite{zhang2020improve}   & 41.5          & 62.2        & 45.1   & 23.5        & 45.0         & 55.3          \\
CWD \cite{shu2021channel}      & 41.7          & 62.0        & 45.5   & 23.3        & 45.5         & 55.5 \\
CWD$^{\sharp}$ \cite{shu2021channel}      & 41.6          & 61.6        & 45.5   & 22.4        & \textbf{45.9}          & 55.0        \\ \hline
ProC-KD (Ours)      & \textbf{42.1}          & \textbf{62.7}        & \textbf{46.0}   & \textbf{23.5}        & 45.8       & \textbf{57.1}           \\

\hline
\end{tabular}
\end{center}
\end{table}

\begin{table*}[t]
\begin{center}
\caption{Ablation study results (\%) of design elements on FoggyCityscapes \cite{sakaridis2018semantic}.}
\label{modules_aba_study}
\begin{tabular}{l|cccccccc|c}
\hline
Method        & bicycle     & bus         & car         &motorcycle & person      & rider       & train       & truck       & mAP \\ \hline
CWD\cite{shu2021channel}           & 53.3        &55.8         & 71.9        &37.5       &57.1         &46.8         &43.2         &50.9         &52.1  \\
ProC-KD (w/proto)      & 53.4        &56.5         & 72.3        &38.1       &55.3         & 47.7        & 45.4        &50.5         &52.4  \\
ProC-KD&\textbf{53.8}&\textbf{57.9}&\textbf{73.1}& \textbf{40.3}&\textbf{57.7}&\textbf{51.2}&\textbf{44.2}&\textbf{51.5}&\textbf{53.7} \\ \hline
\end{tabular}
\end{center}
\end{table*}

\begin{table}[t]\small
\begin{center}
\caption{The mean accuracy (\%) of ProC-KD with the different number of prototypes on the long-tailed CIFAR-100 dataset. }
\begin{tabular}{l|cccc}
\hline
Model/Number & 24    & 48    & 72             & 96 \\ \hline
  ProC-KD     &  77.75 & 78.11 & \textbf{78.32} &78.28    \\ \hline
\end{tabular}
\end{center}
\label{proto_num_aba_study_table}
\end{table}

\subsubsection{Object Detection} We also apply our prototype-guided knowledge distillation method to the standard object detection knowledge distillation task. To make a fair comparison, followed with CWD\cite{shu2021channel} and \cite{zhang2020improve} the teacher model in the experiments is set as Cascade Mask RCNN with ResNeXt101 backbone, and the student model is Faster-RCNN with ResNet-50 backbone. Different from the cross-task knowledge distillation experiments, here, the training dataset of the pre-trained teacher model and knowledge distillation process are both performed on the COCO dataset.

\textbf{Results and Analysis.} Table \ref{stand_obj_det_dist} shows comparison results between our method and other state-of-the-art distillation methods on object detection. The results of other methods are cited from CWD \cite{shu2021channel}. It can be seen that our method outperforms other knowledge distillation methods in different IoU and different object sizes. In particular, we achieve a 1.6\% improvement over the CWD \cite{shu2021channel} on $AP_l$.

Figure \ref{std_pr_curve} shows the error analysis Precision-Recall curve of all objects, large size objects, medium size objects, and small size objects under different conditions on the COCO dataset. The top row is the results of the baseline method CWD \cite{shu2021channel} and the bottom row is the detection results of ProC-KD. We can see that compared with CWD \cite{shu2021channel} our ProC-KD achieves better performance on different IoU thresholds for all different size objects. 
Compared with CWD, Ours improves by 0.029 and 0.024 on large objects and small objects when ignoring the localization errors, respectively. This indicates our method can provide more precise classification information.
Our ProC-KD outperforms the CWD \cite{shu2021channel} by an average of 0.014 on all area objects after ignoring localization errors, ignoring the similar classes from the same supercategory, ignoring all category confusions, and ignoring all false positives, which demonstrate a better location and recognition ability of our method.

\subsection{Ablation Study}

In this part, we ablate the number of prototypes and the important design elements in the proposed prototype-guided knowledge distillation.

\subsubsection{design elements}
We study the effects of the design elements on the FoggyCityscapes. Results are shown in Table {\ref{modules_aba_study}}, it is observed that 
(a) Compared with the CWD \cite{shu2021channel} baseline, 0.3\% (52.4\%-52.1\%) mAP boost can be obtained with the prototype learning module, indicating that the prototype representation is beneficial to knowledge distillation. 
(b) The combination of the prototype learning module and feature augmentation module leads to a significantly mAP improvement, which is 1.3\% (53.7\%-52.4\%). The reason may be that the feature augmentation module enriches the feature which is more related to the object.

\subsubsection{number of prototypes}
We performed ablation experiments on the number of prototypes on the long-tailed CIFAR-100 dataset with ViT model. 
The teacher model is a 12-layer Base version of ViT model, and the student model is a 6-layer ViT model. The number of prototypes is set as 24, 48, 72, and 96 respectively. Here, we only use a different number of prototypes and keep other network settings unchanged to compare the performance of the model.
Table VIII shows that the accuracy of the model increases as the number of prototypes increases, and the best accuracy is 78.32\% when the number of prototypes is 72. It indicates that the small number of prototypes could not learn the generalized representation sufficiently. In our experiment, the number of prototypes in prototype-guided knowledge distillation methods is set to 72.

\begin{table}[!h]
\begin{center}
\caption{Ablation study of loss function hyperparameter on FoggyCityscapes \cite{sakaridis2018semantic}.}
\label{loss_param_aba_study}
\begin{tabular}{c|ccc|c}
\hline
Method/Number        & $ \lambda_{emb}$     & $ \lambda_{pro} $  & $ \lambda_{stu} $   & mAP(\%) \\ \hline
Student  & 0        & 0         & 1              & 35.6  \\
CWD  & 1        & 0         & 1              & 52.1  \\ \hline
1  & 0.3        & 1         & 1              & 49.5  \\
2  & 1        & 0.3         & 1               &52.5  \\
3  & 1        & 1         & 0.3               &51.2  \\
4  & 0.5        & 1         & 1              &50.7  \\
5  & 1        & 0.5         & 1               &52.5  \\
6  & 1        & 1         & 0.5               &52.0  \\ 
7  & 0.8        & 1         & 1              &52.1  \\
8 & 1        & 0.8         & 1               &52.9  \\
9  & 1        & 1         & 0.8               &52.8  \\ 
10  & 1        & 1         & 1               &\textbf{53.7}  \\
 \hline
\end{tabular}
\end{center}
\end{table}

\subsubsection{Hyperparameters}
We conduct the ablation study of hyperparameters in Eq.
~(7)
on the FoggyCityscapes. 
As shown in Table~\ref{loss_param_aba_study}, we set the loss weights $ \lambda_{emb}$, $ \lambda_{pro} $, and $ \lambda_{stu} $ with different values and get the object detection results. 
The ablation study results of loss weights also reveal the effectiveness of our proposed prototype learning method in object detection knowledge distillation.

\section{Conclusion}
To solve the issue of applying a large-scale model to different downstream tasks, we propose a Prototype-guided Cross-task Knowledge Distillation method (ProC-KD), where the label space of the teacher model and the student model is inconsistent. Specifically, the prototype learning module is trained to learn the invariant intrinsic local-level representation with the help of powerful ability from the teacher model. Then, the learned prototypes are used to augment the student model features to improve the generalization ability of the student model. We conduct the experiments in both cross-task knowledge distillation and standard same-task knowledge distillation of image classification and object detection. Both quantitative and qualitative results verify the effectiveness of our proposed method for knowledge distillation.

%
%
%
%
%

\ifCLASSOPTIONcaptionsoff
  \newpage
\fi



\bibliographystyle{IEEEtran}
\bibliography{egbib}
\end{document}